# A Multi-Metric Latent Factor Model for Analyzing High-Dimensional and Sparse data

Di Wu[1,2], Peng Zhang[3], Yi He[4], and Xin Luo[3]

*1 Chongqing Institute of Green and Intelligent Technology, Chinese Academy of Sciences, Chongqing 400714, China.*
*2 Institute of Artificial Intelligence and Blockchain, Guangzhou University, Guangzhou 510006, Guangdong, China.*
*3 School of Computer Science and Technology, Chongqing University of Posts and Telecommunications, Chongqing 400065, China.*
*4 Old Dominion University, Norfolk, Virginia 23462, USA.*
*(e-mail: wudi.cigit@gmail.com)*

**ABSTRACT**

High-dimensional and sparse (HiDS) matrices are omnipresent in a variety of big data-related applications. Latent factor analysis (LFA) is a typical representation learning method that extracts useful yet latent knowledge from HiDS matrices via low-rank approximation. Current LFA-based models mainly focus on a single-metric representation, where the representation strategy designed for the approximation *Loss* function, is fixed and exclusive. However, real-world HiDS matrices are commonly heterogeneous and inclusive and have diverse underlying patterns, such that a single-metric representation is most likely to yield inferior performance. Motivated by this, we in this paper propose a multi-metric latent factor (MMLF) model. Its main idea is two-fold: 1) two vector spaces and three $L_p$-norms are simultaneously employed to develop six variants of LFA model, each of which resides in a unique metric representation space, and 2) all the variants are ensembled with a tailored, self-adaptive weighting strategy. As such, our proposed MMLF enjoys the merits originated from a set of disparate metric spaces all at once, achieving the comprehensive and unbiased representation of HiDS matrices. Theoretical study guarantees that MMLF attains a performance gain. Extensive experiments on eight real-world HiDS datasets, spanning a wide range of industrial and science domains, verify that our MMLF significantly outperforms ten state-of-the-art, shallow and deep counterparts.

## 1 Introduction

Matrices are a norm to describe the pairwise relationships among entities [1, 2]. For example, the user-item matrix has been widely adopted in profiling an e-commerce system [3], with its row and column vectors associating with users and items, respectively. Its each entry indicates the preference or interest of a user towards an item. Indeed, matrices abound in all science and industrial endeavors, influencing practically every aspect of our existence, such as bioinformatics networks [4], wireless sensor networks [5], recommender systems [3], Web service[6], and intelligent transportation [7], to just name a few.

The crux of analyzing these data matrices yielded from large systems lies in their high-dimensional and sparse (HiDS) characteristic [8-14]. This is conceptually and practically tangible, e.g., the numbers of users and items are unmanageably large in nowadays e-commerce platforms such as Amazon or eBay, while the observed entries (user-item interactions) of such large-scaled matrices are relatively very few [3, 15]. As such, the HiDS matrices leave the values of most entries unknown, behind which rich and invaluable knowledge exists [16-21]. Hence, it is of a great significance to learn accurate representations from the HiDS matrices for hidden knowledge extraction [22-27].

Latent factor analysis (LFA) is one of the most popular representation learning methods tailored for HiDS matrices, thanks to its high accuracy, computational efficiency, and ease of scalability [16, 17]. An LFA model employs two low-rank latent factor matrices to produce an approximation of the original HiDS matrix. The two matrices are iteratively trained by minimizing the approximation *Loss* gauged between the originally observed and approximated entries [28], such that the better the observed entries are approximated, the more accurately the missing entries tend to be predicted [8].

Despite their successes, the prior LFA-based models share an essence that their representation strategy, designed for the approximation *Loss* function, is fixed in a priori and exclusive [8]. This leads to a single-metric representation of a target HiDS matrix. However, the real-world HiDS matrices are often heterogeneous and possess complex properties, manifesting the limitations of such a single, fixed, and exclusive representation strategy [8, 29, 30].

In this paper, we proposes a multi-metric latent factor (MMLF) model for accurate representation of the HiDS matrices. Its main idea is two-fold: a) Employing two vector spaces (inner product and distance) and three $L_p$-norms ($L_1$-norm, smooth $L_1$-norm, and $L_2$-norm) to develop six variants of LFA model, each of which owns a unique metric representation space, and b) Ensembling the six metric spaces with a self-tuned, adaptive weighting strategy, where an overall performance gain is theoretically guaranteed. As such, our proposed MMLF model can enjoy the multiple merits originating from each and every metric space, empowering it to significantly outperform the related state-of-the-art models in representing the HiDS matrices.

The main contributions of this paper include:
- An MMLF model is proposed, which strives to make accurate predictions on the missing entries of HiDS matrices by leveraging a multi-metric latent representation space.
- Theoretical analyses are performed to substantiate that our MMLF can aggregate the merits originating from a set of disparate metric spaces, including the inner product space, distance space, $L_1$-norm, smooth $L_1$-norm, and $L_2$-norm.
- Extensive experiments are carried out over eight real-world HiDS datasets that span a wide range of applications, including e-commerce, bioinformatics, and fintech. The results suggest that our MMLF outperforms ten competitors, including both deep and shallow LFA models.

## 2 Related Work

### 2.1 The LFA–based Model

The LFA model is one of the most frequently adopted approaches to represent a HiDS matrix [8]. In the past decade, many advancements of LFA-based models have been proposed from different perspectives [3, 15]. In particular, most of them adopt an $L_2$-norm-oriented Loss, including those are dual-regularization-based [31], generalized non-negative and momentum-based [9], neighborhood-and-location integrated [32], covering-based and neighborhood-aware [33], joint recommendation [34], content features-based [35], confidence-driven [36], and a generalized non-negative based [37], among others. Nevertheless, $L_2$-norm is known to be sensitive to outliers [8]. To counter against this issue, subsequent researchers propose to adopt $L_1$-norm to improve an LFA model's robustness to outliers [8, 38, 39]. However, all the above models are developed on inner product space that violates the triangle inequality [29, 30], making them prone to ignoring the fine-grained similarity (i.e., local similarity). Alternatively, Hsieh et al. propose to apply metric learning [30] to recommender systems, which can capture the locally user-wise or item-wise similarities represented by Enclidean distance directly. On this foundation, Zhang et al. investigate the metric vector space (i.e., distance vector space) to develop an LFA model for recommendation tasks [29].

Notably, all the above LFA-based models are developed with a single-metric representation strategy to HiDS data. Recall Example 1, however, one metric representation strategy commonly has its unique limitations. As a result, they cannot comprehensively represent a target HiDS matrix's characteristics. In comparison, MMLF adopts the multi-metric representation strategy by aggregating the multi-merits of inner product space, distance space, $L_1$-norm, smooth $L_1$-norm, and $L_2$-norm together, which makes it achieve highly accurate representation of a HiDS matrix.

### 2.2 The Deep Learning-based Model

Recently, deep neural networks (DNNs) [40, 41] have attracted extensive attention in analyzing HiDS data from recommender systems due to their powerful representation learning ability [42]. Zhang et al. provide a comprehensive review on DNNs-based recommender systems [3]. To date, various sophisticated DNNs-based models have been proposed. Representative studies include autoencoder-based [43], variational autoencoder-based [44], neural collaborative filtering-based [42], neural factorization-based [45], federated meta-learning-based [46], neural rating regression-based [47], deep-rating and review-neural-network-based [48], and latent relevant metric learning-based [49] models. Moreover, with the development of graph neural networks (GNNs) [50-52], some researchers also propose to employ GNNs to analyze HiDS data from recommender systems, including graph autoencoder-based [53], inductive graph-based [54], multi-component graph-based [55], neural graph collaborative filtering-based [56, 57], and self-supervised graph-based [58] methods.

Compared with the above deep learning-based models, MMLF possesses its significance in the following aspects: 1) MMLF has much higher computational efficiency because it is trained only on observed entries of an input HiDS matrix; 2) S. Rendle et al. demonstrate that the LFA model is still highly competitive with them [59]. MMLF is developed by improving LFA with a multi-metric representation strategy. For these reasons, we advocate that our proposed MMLF is conceptual comparable with the DNN-based LFA models, while craving much less input information.

## 3 The Proposed MMLF Model

### 3.1 Building Base Models

The *Loss* function (1) is the core of an LFA model. We pairwise employ the two vector spaces and the three $L_p$-norms to design the functions of $l(\Delta_{m,n})$ and $\hat{h}_{m,n}(p_m,q_n)$ to build six base models, as shown in Table 1. Besides, we incorporate the training bias into their objective function to further improve their representation learning ability [8, 16]. Next, we detail them.

Table 1. The Loss function of six base models

| Base model | $\hat{h}_{m,n}(p_m,q_n)$ | $l(\Delta_{m,n})$ | Characteristic |
|---|---|---|---|
| MMLF-1 | $p_m q_n$ | $|\Delta_{m,n}|$ | Global similarity & Robustness |
| MMLF-2 | $p_m q_n$ | $(\Delta_{m,n})^2$ | Global similarity & Stablity |
| MMLF-3 | $p_m q_n$ | $\|\Delta_{m,n}\|, if |\Delta_{m,n}|>1$ <br> $(\Delta_{m,n})^2, if |\Delta_{m,n}|\leq 1$ | Global similarity & Robustness/Stablity |
| MMLF-4 | $\|p_m-q_n\|_2$ | $|\Delta_{m,n}|$ | Local similarity & Robustness |
| MMLF-5 | $\|p_m-q_n\|_2$ | $(\Delta_{m,n})^2$ | Local similarity & Stablity |
| MMLF-6 | $\|p_m-q_n\|_2$ | $|\Delta_{m,n}|, if |\Delta_{m,n}|>1$ <br> $(\Delta_{m,n})^2, if |\Delta_{m,n}|\leq 1$ | Local similarity & Robustness/Stablity |

$\|\cdot\|_2$ denotes the $L_2$-norm of a vector.

### 3.1.1 MMLF-1 (inner product space and $L_1$-norm)

Following Table 1, we design the objective function of MMLF-1 with training bias as follows:

$$\varepsilon\left(P^{k=1}, Q^{k=1}\right) = \sum_{h_{m,n} \in H_O} \left| h_{m,n} - \underbrace{p_m^{k=1} q_n^{k=1}}_{\text{Inner product space}} - \underbrace{\left(b_m^{k=1} + b_n^{k=1}\right)}_{\text{Training bias}} \right| + \frac{1}{2}\lambda \underbrace{\sum_{h_{m,n} \in H_O} \left(\left(p_m^{k=1}\right)^2 + \left(q_n^{k=1}\right)^2 + \left(b_m^{k=1}\right)^2 + \left(b_n^{k=1}\right)^2\right)}_{\text{Regularization}}. \quad (1)$$

$L_1$-norm oriented loss

We adopt SGD to minimize (1) to obtain the training rules of $P^{k=1}$ and $Q^{k=1}$ of MMLF-1 as follows:

$$\text{for } h_{m,n} \in H_O : \begin{cases} \Delta_{m,n}^{k=1} \geq 0 : \begin{cases} p_m^{k=1} \leftarrow (1-\eta\lambda) p_m^{k=1} + \eta q_n^{k=1} \\ q_n^{k=1} \leftarrow (1-\eta\lambda) q_n^{k=1} + \eta p_m^{k=1} \\ b_m^{k=1} \leftarrow (1-\eta\lambda) b_m^{k=1} + \eta \\ b_n^{k=1} \leftarrow (1-\eta\lambda) b_n^{k=1} + \eta \end{cases} \\ \Delta_{m,n}^{k=1} < 0 : \begin{cases} p_m^{k=1} \leftarrow (1-\eta\lambda) p_m^{k=1} - \eta q_n^{k=1} \\ q_n^{k=1} \leftarrow (1-\eta\lambda) q_n^{k=1} - \eta p_m^{k=1} \\ b_m^{k=1} \leftarrow (1-\eta\lambda) b_m^{k=1} - \eta \\ b_n^{k=1} \leftarrow (1-\eta\lambda) b_n^{k=1} - \eta \end{cases} \end{cases}, \quad (2)$$

where $\Delta_{m,n}^{k=1} = h_{m,n} - p_m^{k=1} q_n^{k=1} - b_m^{k=1} - b_n^{k=1}$.

### 3.1.2 MMLF-2 (inner product space and $L_2$-norm)

Following Table 1, we design the objective function of MMLF-2 with training bias as follows:

$$\varepsilon\left(P^{k=2}, Q^{k=2}\right) = \frac{1}{2}\sum_{h_{m,n} \in H_O} \underbrace{\left(h_{m,n} - p_m^{k=2} q_n^{k=2} - b_m^{k=2} - b_n^{k=2}\right)^2}_{L_2\text{-norm oriented loss}} + \frac{1}{2}\lambda \sum_{h_{m,n} \in H_O} \left(\left(p_m^{k=2}\right)^2 + \left(q_n^{k=2}\right)^2 + \left(b_m^{k=2}\right)^2 + \left(b_n^{k=2}\right)^2\right). \quad (3)$$

We adopt SGD to minimize (3) to obtain the training rules of $P^{k=2}$ and $Q^{k=2}$ of MMLF-2 as follows:

$$\text{for } h_{m,n} \in H_O : \begin{cases} p_m^{k=2} \leftarrow (1-\eta\lambda) p_m^{k=2} + \eta \Delta_{m,n}^{k=2} q_n^{k=2} \\ q_n^{k=2} \leftarrow (1-\eta\lambda) q_n^{k=2} + \eta \Delta_{m,n}^{k=2} p_m^{k=2} \\ b_m^{k=2} \leftarrow (1-\eta\lambda) b_m^{k=2} + \eta \Delta_{m,n}^{k=2} \\ b_n^{k=2} \leftarrow (1-\eta\lambda) b_n^{k=2} + \eta \Delta_{m,n}^{k=2} \end{cases}, \quad (4)$$

where $\Delta_{m,n}^{k=2} = h_{m,n} - p_m^{k=2} q_n^{k=2} - b_m^{k=2} - b_n^{k=2}$.

### 3.1.3 MMLF-3 (inner product space and smooth $L_1$-norm)

Following Table 1, we design the objective function of MMLF-3 with training bias as follows:

$$\varepsilon\left(P^{k=3}, Q^{k=3}\right) = \sum_{h_{m,n} \in H_O} \underbrace{\begin{pmatrix} \left|\Delta_{m,n}^{k=3}\right|, & if \left|\Delta_{m,n}^{k=3}\right|>1 \\ \left(\Delta_{m,n}^{k=3}\right)^2, & if \left|\Delta_{m,n}^{k=3}\right|\leq 1 \end{pmatrix}}_{\text{Smooth } L_1\text{-norm oriented loss}} + \frac{1}{2}\lambda \sum_{h_{m,n} \in H_O} \left(\left(p_m^{k=3}\right)^2 + \left(q_n^{k=3}\right)^2 + \left(b_m^{k=3}\right)^2 + \left(b_n^{k=3}\right)^2\right), \quad (5)$$

where $\Delta_{m,n}^{k=3} = h_{m,n} - p_m^{k=3} q_n^{k=3} - b_m^{k=3} - b_n^{k=3}$. We adopt SGD to minimize (5) to obtain the training rules of $P^{k=3}$ and $Q^{k=3}$ of MMLF-3 as follows:

$$\text{for } h_{m,n} \in H_O : \begin{cases} \Delta_{m,n}^{k=3} > 1 : \begin{cases} p_m^{k=3} \leftarrow (1-\eta\lambda) p_m^{k=3} + \eta q_n^{k=3} \\ q_n^{k=3} \leftarrow (1-\eta\lambda) q_n^{k=3} + \eta p_m^{k=3} \\ b_m^{k=3} \leftarrow (1-\eta\lambda) b_m^{k=3} + \eta \\ b_n^{k=1} \leftarrow (1-\eta\lambda) b_n^{k=3} + \eta \end{cases} \\ \Delta_{m,n}^{k=3} < -1 : \begin{cases} p_m^{k=3} \leftarrow (1-\eta\lambda) p_m^{k=3} - \eta q_n^{k=3} \\ q_n^{k=3} \leftarrow (1-\eta\lambda) q_n^{k=3} - \eta p_m^{k=3} \\ b_m^{k=3} \leftarrow (1-\eta\lambda) b_m^{k=3} - \eta \\ b_n^{k=3} \leftarrow (1-\eta\lambda) b_n^{k=3} - \eta \end{cases} \\ |\Delta_{m,n}^{k=3}| \leq 1 : \begin{cases} p_m^{k=3} \leftarrow (1-\eta\lambda) p_m^{k=3} + 2\eta\Delta_{m,n}^{k=3} q_n^{k=3} \\ q_n^{k=3} \leftarrow (1-\eta\lambda) q_n^{k=3} + 2\eta\Delta_{m,n}^{k=3} p_m^{k=3} \\ b_m^{k=3} \leftarrow (1-\eta\lambda) b_m^{k=3} + 2\eta\Delta_{m,n}^{k=3} \\ b_n^{k=3} \leftarrow (1-\eta\lambda) b_n^{k=3} + 2\eta\Delta_{m,n}^{k=3} \end{cases} \end{cases}. \quad (6)$$

### 3.1.4 MMLF-4 (distance space and $L_1$-norm)

Following Table 1, we design the objective function of MMLF-4 with training bias as follows:

$$\varepsilon(P^{k=4}, Q^{k=4}) = \sum_{h_{m,n} \in H_O} \left| h_{m,n} - \underbrace{\|p_m^{k=4} - q_n^{k=4}\|_2}_{\text{Distance space}} - b_m^{k=4} - b_n^{k=4} \right| + \frac{1}{2}\lambda \sum_{h_{m,n} \in H_O} \left( (p_m^{k=4})^2 + (q_n^{k=4})^2 + (b_m^{k=4})^2 + (b_n^{k=4})^2 \right). \quad (7)$$

We adopt SGD to minimize (7) to obtain the training rules of $P^{k=4}$ and $Q^{k=4}$ of MMLF-4 as follows:

$$\text{for } h_{m,n} \in H_O : \begin{cases} \Delta_{m,n}^{k=4} \geq 0 : \begin{cases} p_m^{k=4} \leftarrow (1-\eta\lambda) p_m^{k=4} + \eta \frac{p_m^{k=4} - q_n^{k=4}}{\|p_m^{k=4} - q_n^{k=4}\|_2} \\ q_n^{k=4} \leftarrow (1-\eta\lambda) q_n^{k=4} - \eta \frac{p_m^{k=4} - q_n^{k=4}}{\|p_m^{k=4} - q_n^{k=4}\|_2} \\ b_m^{k=4} \leftarrow (1-\eta\lambda) b_m^{k=4} + \eta \\ b_n^{k=4} \leftarrow (1-\eta\lambda) b_n^{k=4} + \eta \end{cases} \\ \Delta_{m,n}^{k=4} < 0 : \begin{cases} p_m^{k=4} \leftarrow (1-\eta\lambda) p_m^{k=4} - \eta \frac{p_m^{k=4} - q_n^{k=4}}{\|p_m^{k=4} - q_n^{k=4}\|_2} \\ q_n^{k=4} \leftarrow (1-\eta\lambda) q_n^{k=4} + \eta \frac{p_m^{k=4} - q_n^{k=4}}{\|p_m^{k=4} - q_n^{k=4}\|_2} \\ b_m^{k=4} \leftarrow (1-\eta\lambda) b_m^{k=4} - \eta \\ b_n^{k=4} \leftarrow (1-\eta\lambda) b_n^{k=4} - \eta \end{cases} \end{cases}, \quad (8)$$

where $\Delta_{m,n}^{k=4} = h_{m,n} - \|p_m^{k=4} - q_n^{k=4}\|_2 - b_m^{k=4} - b_n^{k=4}$.

### 3.1.5 MMLF-5 (distance space and $L_2$-norm)

Following Table 1, we design the objective function of MMLF-5 with training bias as follows:

$$\varepsilon(P^{k=4}, Q^{k=4}) = \frac{1}{2} \sum_{h_{m,n} \in H_O} \left( h_{m,n} - \|p_m^{k=4} - q_n^{k=4}\|_2 - b_m^{k=4} - b_n^{k=4} \right)^2 + \frac{1}{2}\lambda \sum_{h_{m,n} \in H_O} \left( (p_m^{k=4})^2 + (q_n^{k=4})^2 + (b_m^{k=4})^2 + (b_n^{k=4})^2 \right). \quad (9)$$

We adopt SGD to minimize (9) to obtain the training rules of $P^{k=5}$ and $Q^{k=5}$ of MMLF-5 as follows:

$$\text{for } h_{m,n} \in H_O : \begin{cases} p_m^{k=5} \leftarrow (1-\eta\lambda) p_m^{k=5} + \eta\Delta_{m,n}^{k=5} \frac{p_m^{k=5} - q_n^{k=5}}{\|p_m^{k=5} - q_n^{k=5}\|_2} \\ q_n^{k=5} \leftarrow (1-\eta\lambda) q_n^{k=5} - \eta\Delta_{m,n}^{k=5} \frac{p_m^{k=5} - q_n^{k=5}}{\|p_m^{k=5} - q_n^{k=5}\|_2} \\ b_m^{k=5} \leftarrow (1-\eta\lambda) b_m^{k=5} + \eta\Delta_{m,n}^{k=5} \\ b_n^{k=5} \leftarrow (1-\eta\lambda) b_n^{k=5} + \eta\Delta_{m,n}^{k=5} \end{cases}, \quad (10)$$

where $\Delta_{m,n}^{k=5} = h_{m,n} - \|p_m^{k=5} - q_n^{k=5}\|_2 - b_m^{k=5} - b_n^{k=5}$.

*3.1.6 MMLF-6 (distance space and smooth L₁-norm)*

Following Table 1, we design the objective function of MMLF-6 with training bias as follows:

$$\varepsilon\left(P^{k=6}, Q^{k=6}\right) = \sum_{h_{m,n} \in H_O} \begin{pmatrix} \left|\Delta_{m,n}^{k=6}\right|, & if \ \left|\Delta_{m,n}^{k=6}\right| > 1 \\ \left(\Delta_{m,n}^{k=6}\right)^2, & if \ \left|\Delta_{m,n}^{k=6}\right| \leq 1 \end{pmatrix} + \frac{1}{2}\lambda \sum_{h_{m,n} \in H_O} \left(\left(p_m^{k=6}\right)^2 + \left(q_n^{k=6}\right)^2 + \left(b_m^{k=6}\right)^2 + \left(b_n^{k=6}\right)^2\right), \quad (11)$$

where $\Delta_{m,n}^{k=6} = h_{m,n} - \|p_m^{k=6} - q_n^{k=6}\|_2 - b_m^{k6} - b_n^{k=6}$. We adopt SGD to minimize (11) to obtain the training rules of $P^{k=6}$ and $Q^{k=6}$ of MMLF-6 as follows:

$$\text{for } h_{m,n} \in H_O: \begin{cases} \Delta_{m,n}^{k=6} > 1: \begin{cases} p_m^{k=6} \leftarrow (1-\eta\lambda)p_m^{k=6} + \eta \frac{p_m^{k=6} - q_n^{k=6}}{\|p_m^{k=6} - q_n^{k=6}\|_2} \\ q_n^{k=6} \leftarrow (1-\eta\lambda)q_n^{k=6} - \eta \frac{p_m^{k=6} - q_n^{k=6}}{\|p_m^{k=6} - q_n^{k=6}\|_2} \\ b_m^{k=6} \leftarrow (1-\eta\lambda)b_m^{k=6} + \eta \\ b_n^{k=6} \leftarrow (1-\eta\lambda)b_n^{k=6} + \eta \end{cases} \\ \Delta_{m,n}^{k=6} < -1: \begin{cases} p_m^{k=6} \leftarrow (1-\eta\lambda)p_m^{k=6} - \eta \frac{p_m^{k=6} - q_n^{k=6}}{\|p_m^{k=6} - q_n^{k=6}\|_2} \\ q_n^{k=6} \leftarrow (1-\eta\lambda)q_n^{k=6} + \eta \frac{p_m^{k=6} - q_n^{k=6}}{\|p_m^{k=6} - q_n^{k=6}\|_2} \\ b_m^{k=6} \leftarrow (1-\eta\lambda)b_m^{k=6} - \eta \\ b_n^{k=6} \leftarrow (1-\eta\lambda)b_n^{k=6} - \eta \end{cases} \\ \left|\Delta_{m,n}^{k=6}\right| \leq 1: \begin{cases} p_m^{k=6} \leftarrow (1-\eta\lambda)p_m^{k=6} + 2\eta\Delta_{m,n}^{k=6} \frac{p_m^{k=6} - q_n^{k=6}}{\|p_m^{k=6} - q_n^{k=6}\|_2} \\ q_n^{k=6} \leftarrow (1-\eta\lambda)q_n^{k=6} - 2\eta\Delta_{m,n}^{k=6} \frac{p_m^{k=6} - q_n^{k=6}}{\|p_m^{k=6} - q_n^{k=6}\|_2} \\ b_m^{k=6} \leftarrow (1-\eta\lambda)b_m^{k=6} + 2\eta\Delta_{m,n}^{k=6} \\ b_n^{k=6} \leftarrow (1-\eta\lambda)b_n^{k=6} + 2\eta\Delta_{m,n}^{k=6} \end{cases} \end{cases} \quad (12)$$

### 3.2 Self-ensembling

The ensemble is a great way to aggregate multi-models. It requires the base models are diversified and accurate [60, 61]. In MMLF, the six base models are built by the different two vector spaces and three $L_p$-norms, which guarantees their diversity. Besides, they are the variant of an LFA model that has been demonstrated to be accurate, which guarantees their accuracy. Hence, they satisfy the two requirements of the ensemble. To finely ensemble them, we make their weights be self-adaptive according to the training *loss*. The main idea is to increase *k*-th base model's weight if its partial training *loss* decreases at *t*-th training iteration, and decrease its weight otherwise. For theoretically validating the effectiveness of this strategy, we firstly present the following definitions.

**Definition 1.** Let $Pl^k(t)$ be the partial training *loss* of *k*-th base model at *t*-th training iteration, then it can be computed as follows:

$$Pl^k(t) = \sum_{h_{m,n} \in H_O} \left|h_{m,n} - \hat{h}_{m,n}^k\right|, \hat{h}_{m,n}^k = p_m^k q_n^k + b_m^k + b_n^k, \quad s.t. \ k = 1,2,3,$$

$$\hat{h}_{m,n}^k = \|p_m^k - q_n^k\|_2 + b_m^k + b_n^k, \ s.t. \ k = 4,5,6. \quad (13)$$

**Definition 2.** Let $Cl^k(t)$ be the cumulative training *loss* of $Pl^k(t)$ until *t*-th training iteration, then it can be computed as follows:

$$Cl^k(t) = \sum_{i=1}^{t} Pl^k(t). \quad (14)$$

**Definition 3.** Let $\alpha^k(t)$ be the ensemble weight of *k*-th base model at *t*-th training iteration, then it can be set as follows:

$$\alpha^k(t) = \frac{e^{-\zeta Cl^k(t)}}{\sum_{k=1}^{6} e^{-\zeta Cl^k(t)}}, \quad (15)$$

where $\zeta$ is a balance coefficient controlling the ensemble of base models during the training processes.

Then, based on definitions 3–5, the final predictions of MMLF at *t*-th training iteration are obtained as follows:

$$\hat{h}_{m,n} = \sum_{k=1}^{6} \alpha^k(t)\hat{h}_{m,n}^k \quad (16)$$

## 4 Experiments

### 4.1 General Settings

**Datasets.** Four frequently used HiDS datasets are selected to conduct the experiments. They are real datasets generated from different fields, including e-commerce, bioinformatics, and fintech. Table 2 summarizes their details.

Table 2. Properties of all the datasets.

| No. | Name | $|M|$ | $|N|$ | $|H_O|$ | Density* |
|---|---|---|---|---|---|
| D1 | MovieLens_20M | 138,493 | 26,744 | 20,000,263 | 0.54% |
| D2 | MovieLens_10M | 71,567 | 65,133 | 10,000,054 | 0.21% |
| D3 | Dating | 135,359 | 168,791 | 17,359,346 | 0.08% |
| D4 | EachMovie | 72,916 | 1628 | 2,811,718 | 2.37% |

*Density denotes the percentage of observed entries in the HiDS matrix.

**Evaluation Metrics.** Missing data prediction is an important and crucial issue in evaluating the representation of a HiDS matrix [62-68]. To evaluate the accuracy of missing data prediction, we adopt the widely used root mean squared error (RMSE) and mean absolute error (MAE) as the evaluation metrics [3, 8, 9, 15, 16]. They are calculated by:

$$RMSE = \sqrt{\left(\sum_{h_{m,n} \in \Gamma} \left(h_{m,n} - \hat{h}_{m,n}\right)^2\right) \bigg/ |\Gamma|}, \quad MAE = \left(\sum_{h_{m,n} \in \Gamma} \left|h_{m,n} - \hat{h}_{m,n}\right|\right) \bigg/ |\Gamma|$$

where $\Gamma$ denotes the testing set.

**Baselines.** We compare the proposed MMLF model with nine related state-of-the-art models, including five LFA-based models (BLF, SL-LF, L$^3$F, GFNLF, and FML) and five deep learning-based models (AutoRec, NRR, MetaMF, and NeuMF).

### 4.2 Performance Comparison

Table 3 records the prediction accuracy of all the involved models on different datasets. From Table 3, we find that MMLF achieves the lowest RMSE/MAE on most cases. Therefore, the results demonstrate that MMLF significantly outperforms the comparison models in terms of accuracy of predicting the missing data of a HiDS matrix.

Table 3. The comparison results on prediction accuracy.

| Dataset | Metric | BLF | SL-LF | L$^3$F | GFNLF | FML | AutoRec | NRR | MetaMF | NeuMF | MMLF |
|---|---|---|---|---|---|---|---|---|---|---|---|
| D1 | RMSE | 0.7737 | 0.7790 | 0.7806 | 0.7792 | 0.7802 | 0.7802 | 0.7798 | 0.7905 | 0.7931 | 0.7693 |
| D1 | MAE | 0.5886 | 0.5857 | 0.5863 | 0.5901 | 0.5905 | 0.5947 | 0.6018 | 0.6221 | 0.6049 | 0.5789 |
| D2 | RMSE | 0.7819 | 0.7887 | 0.7899 | 0.7872 | 0.7892 | 0.7865 | 0.7834 | 0.8010 | 0.7969 | 0.7787 |
| D2 | MAE | 0.5999 | 0.5980 | 0.5981 | 0.6032 | 0.6037 | 0.6048 | 0.6035 | 0.6169 | 0.6118 | 0.5909 |
| D3 | RMSE | 1.8066 | 1.7829 | 1.7704 | 1.8124 | 1.8012 | 1.8027 | 1.8101 | 1.8013 | 1.8257 | 1.7591 |
| D3 | MAE | 1.2392 | 1.1686 | 1.1781 | 1.2442 | 1.2412 | 1.2610 | 1.2303 | 1.2406 | 1.2635 | 1.1440 |
| D4 | RMSE | 0.2251 | 0.2260 | 0.2256 | 0.2272 | 0.2258 | 0.2305 | 0.2301 | 0.2365 | 0.2361 | 0.2212 |
| D4 | MAE | 0.1732 | 0.1729 | 0.1698 | 0.1774 | 0.1713 | 0.1784 | 0.1774 | 0.1813 | 0.1882 | 0.1683 |

## 5 Conclusion

This paper proposes a multi-metric latent factor (MMLF) model for highly accurate representation of high-dimensional and sparse (HiDS) matrices. Its main idea is three-fold: 1) employing two vector spaces and three $L_p$-norms to develop six variants of latent factor analysis (LFA) model, where each one has a different metric representation strategy, and 2) ensembling these variants with a carefully designed adaptive weight strategy. Theoretical and empirical studies show that MMLF can self-adaptively aggregate the multi-merits originated from the two vector spaces and the three $L_p$-norms. Experimental results on eight real datasets collected from a variety of science and industrial domains demonstrate that MMLF has significantly higher accuracy in predicting the missing data of a HiDS matrix. In the future, we plan to incorporate more metrics, e.g., $L_{2,1}$-norm, into the metric space of our MMLF model.